%% file: main.tex
\documentclass{article}
\usepackage{iclr2023_conference,times}

\input{math_commands}

\usepackage[colorlinks=true,citecolor=blue]{hyperref}
\usepackage{url}
\usepackage{wrapfig}
\usepackage{graphicx}
\usepackage{amssymb}
\usepackage{lipsum}
\usepackage{enumitem}
\usepackage{multirow}

\title{ChiroDiff: Modelling chirographic data with Diffusion Models}

\author{Ayan Das\textsuperscript{1,2}, Yongxin Yang\textsuperscript{1,3}, Timothy Hospedales\textsuperscript{1,4,5}, Tao Xiang\textsuperscript{1,2} \& Yi-Zhe Song\textsuperscript{1,2} \vspace*{.15cm} \\ 
\textsuperscript{1}SketchX, CVSSP, University of Surrey;
\textsuperscript{2}iFlyTek-Surrey Joint Research Centre on AI; \\ \textsuperscript{3}Queen Mary University of London; \textsuperscript{4}University of Edinburgh, 
\textsuperscript{5}Samsung AI Centre, Cambridge \vspace{0.1cm}\\
a.das@surrey.ac.uk,\ yongxin.yang@qmul.ac.uk,\ t.hospedales@ed.ac.uk,\ t.xiang@surrey.ac.uk \\ y.song@surrey.ac.uk
}

\newcommand{\modified}[1]{#1}
\newcommand{\keypoint}[1]{\vspace{0.5em}\noindent\textbf{#1}\quad}
\iclrfinalcopy

\begin{document}

\maketitle

\begin{abstract}
Generative modelling over continuous-time geometric constructs, a.k.a \emph{chirographic data} such as handwriting, sketches, drawings etc., have been accomplished through autoregressive distributions. Such strictly-ordered discrete factorization however falls short of capturing key properties of chirographic data -- it fails to build holistic understanding of the temporal concept due to one-way visibility (causality). Consequently, temporal data has been modelled as discrete token sequences of fixed sampling rate instead of capturing the true underlying concept. In this paper, we introduce a powerful model-class namely \emph{Denoising Diffusion Probabilistic Models} or DDPMs for chirographic data that specifically addresses these flaws. Our model named ``\textsc{ChiroDiff}'', being non-autoregressive, learns to capture holistic concepts and therefore remains resilient to higher temporal sampling rate up to a good extent. Moreover, we show that many important downstream utilities (e.g. conditional sampling, creative mixing) can be flexibly implemented using \textsc{ChiroDiff}. We further show some unique use-cases like stochastic vectorization, de-noising/healing, abstraction are also possible with this model-class. We perform quantitative and qualitative evaluation of our framework on relevant datasets and found it to be better or on par with competing approaches. Please visit our project page for more details: \href{https://ayandas.me/chirodiff}{https://ayandas.me/chirodiff}.
\end{abstract}

\vspace*{-0.5cm}

\section{Introduction} \label{sec:intro}

\begin{figure}[b]
    \centering
    \includegraphics[width=0.9\linewidth]{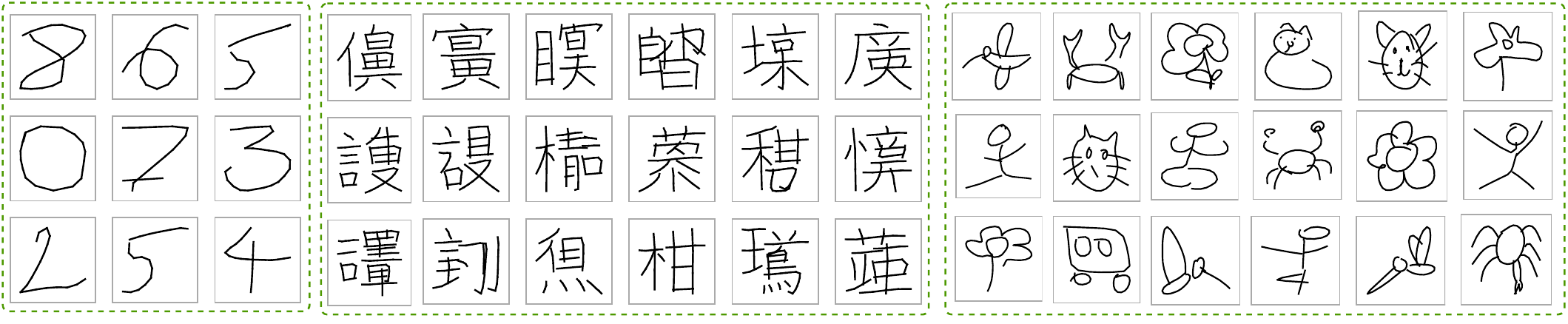}
    \vspace*{-0.3cm}
    \caption{Unconditional samples from \textsc{ChiroDiff} trained on VMNIST, KanjiVG and \emph{Quick, Draw!}.}
    \label{fig:banner}
\end{figure}

%% general stuff about sketches etc, other tasks and making the case for generative modelling
Chirographic data like handwriting, sketches, drawings etc. are ubiquitous in modern day digital contents, thanks to the widespread adoption of touch screen and other interactive devices (e.g. AR/VR sets). While supervised downstream tasks on such data like sketch-based image retrieval (SBIR) \citep{liu2020scenesketcher,kaiyue19sbir}, semantic segmentation \citep{yang2021sketchgnn,wang2020multi}, classification \citep{sketchanet1,sketchanet2} continue to flourish due to higher commercial demand, unsupervised generative modelling remains slightly under-explored. Recently however, with the advent of large-scale datasets, generative modelling of chirographic data started to gain traction. Specifically, models have been trained on generic doodles/drawings data \citep{ha2017neural}, or more ``specialized'' entities like fonts \citep{fontgen_iccv}, diagrams \citep{dididataset,aksan2020cose}, SVG Icons \citep{carlier2020deepsvg} etc. Building unconditional neural generative models not only allows understanding the distribution of chirographic data but also enables further downstream tasks (e.g. segmentation, translation) by means of conditioning.

\begin{figure}
    \centering
    \includegraphics[width=0.9\linewidth]{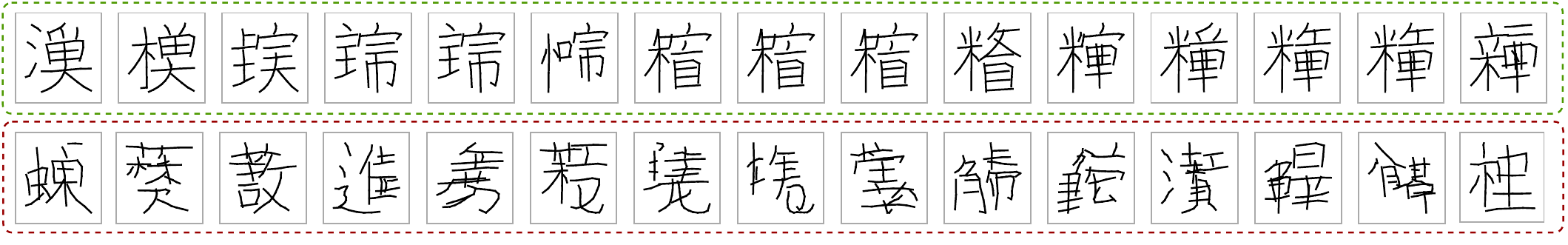}
    \vspace{-0.4cm}
    \caption{Latent space interpolation (Top) with \textsc{ChiroDiff} using DDIM sampler and (Bottom) with auto-regressive model. \textsc{ChiroDiff}'s latent space is much more effective with compositional structures for complex data.}
    \label{fig:holistic_interp}
    \vspace{-0.6cm}
\end{figure}

%% clarify what is Creative & Perceptive model, and their benefits
By far, learning neural models over continuous-time chirographic structures have been facilitated broadly by two different representations -- grid-based raster image and vector graphics. Raster format, the de-facto representation for natural images, has served as an obvious choice for chirographic structures \citep{sketchanet1,sketchanet2}. The static nature of the representation however does not provide the means for modelling the underlying \emph{creative process} that is inherent in drawing. ``Creative models'', powered by topology specific vector formats \citep{carlier2020deepsvg,aksan2020cose,ha2017neural,fontgen_iccv,das2022sketchode}, on the other hand, are specifically motivated to mimic this dynamic creation process. They build distributions of a chirographic entity (e.g., a sketch) $X$ with a specific topology (drawing direction, stroke order etc), i.e. $p_{\theta}(X)$. 
% The availability of large-scale vectorized datasets \citep{dididataset,ha2017neural,carlier2020deepsvg} also helped creative models to gain traction and showed popularity among digital content creators. 
% These class of models aptly exploit the inherent sparsity present in chirographic data.
Majority of the creative models are designed with autoregressive distributions \citep{ha2017neural,aksan2020cose,sketchformer}. Such design choice is primarily due to vector formats having variable lengths, which is elegantly handled by autoregression. Doing so, however, restrict the model from gaining full visibility of the data and fails to build holistic understanding of the temporal concepts. A simple demonstration of its latent-space interpolation confirms this hypothesis (Figure~\ref{fig:holistic_interp}). The other possibility is to drop the ordering/sequentiality of the points entirely and treat chirographic data as 2D point-sets and use prominent techniques from 3D point-cloud modelling \citep{shitong_denoising_score,shitong_diff3d,shapegf}. However, point-set representation does not fit chirographic data well due to its inherently unstructured nature. In this paper, with \textsc{ChiroDiff}, we find a sweet spot and propose a framework that uses non-autoregressive density while retaining its sequential nature.

% The superscript $\pi$ simply denotes a the specific ordering enforced by the sequence. We will drop the $\pi$ notation from now onward unless necessary.
% \item \textbf{Topology-agnostic $\mathcal{X}$ for \emph{perception} model:}\quad Instead of traditional raster images, we train perceptive models with \emph{point-sets}. We define $\mathcal{X} = \{ \cdots, \mathbf{x}^{(j)}, \cdots \}$ where each $\mathbf{x}^{(j)} \in \mathbb{R}^2$. By using a set representation, we take advantage of the inherent sparsity present in chirographic structures while keeping it agnostic to topology. This approach is mostly used in modelling 3D shapes \citep{pointnet,pointgrow,pointflow}. Please note that timestamps are irrelevant for this representation, and for any ordering $\pi_1, \pi_2 \in \Pi$, we have $\mathcal{X}^{\pi_1} \equiv \mathcal{X}^{\pi_2}$.

%% arbitrary resolution
Another factor in traditional neural chirographic models that limit the representation is effective handling of temporal resolution. Chirographic structures are inherently continuous-time entities as rightly noted by \citet{das2022sketchode}. Prior works like SketchRNN \citep{ha2017neural} modelled continuous-time chirographic data as discrete token sequence or motor program. Due to limited visibility, these models do not have means to accommodate different sampling rates and are therefore specialized to learn for one specific temporal resolution (seen during training), leading to the loss of spatial/temporal scalability essential for digital contents. Even though there have been attempts \citep{aksan2020cose,das2020bziersketch} to directly represent continuous-time entities with their underlying geometric parameters, most of them still possess some form of autoregression. Recently, SketchODE \citep{das2022sketchode} approached to solve this problem by using Neural ODE (abbreviated as NODE) \citep{neuralode} for representing time-derivative of continuous-time functions. However, the computationally restrictive nature of NODE's training algorithm makes it extremely hard to train and adopt beyond simple temporal structures. \textsc{ChiroDiff}, having visibility of the entire sequence, is capable of implicitly modelling the sampling rate from data and consequently is robust to learning the continuous-time temporal concept that underlies the discrete motor program. In that regard, \textsc{ChiroDiff} outperforms \citet{das2022sketchode} significantly by adopting a model-class superior in terms of computational costs and representational power while training on similar data.

We chose \emph{Denoising Diffusion Probabilstic Models} (abbr. as DDPMs) as the model class due to their spectacular ability to capture both diversity and fidelity \citep{dalle_openai,glide_openai}. Furthermore, Diffusion Models are gaining significant popularity and nearly replacing GANs in wide range of visual synthesis tasks due to their stable training dynamics and generation quality. A surprising majority of existing works on Diffusion Model is solely based or specialized to grid-based raster images, leaving important modalities like sequences behind. Even though there are some isolated works on modelling sequential data, but they have mostly been treated as fixed-length entities \citep{csdi_imputation}. Our proposed model, in that regard, is \modified{one of} the first \modified{models} to exhibit the potential to apply Diffusion Model on continuous-time entities. To this end, our generative model generates $X$ by transforming a discretized \emph{brownian motion} with unit step size.

% Our \emph{perception} model is also one of it's kind in terms of data modality. We employ diffusion models to capture distribution over set-structured chirographic data. Even though there have some works \citep{shitong_diff3d,shapegf} on modelling point-sets in 3D, our usage is fundamentally different. \citet{shitong_diff3d} and \citet{shapegf} uses Diffusion Model for encoding point-sets into explicit geometric shape parameters $\Theta = G(X)$ followed by leaning a generative model $p(\Theta)$ (VAE or GAN) on the parameters. Our \emph{perception} model leverages Diffusion Model to directly define a generative model on the set-space $X$ without building explicit representation.

%% a little more details about the downstream applications
We consider learning stochastic generative model for continuous-time chirographic data both in unconditional (samples shown in Figure~\ref{fig:banner}) and conditional manner.
%TH Note: SketchODE had it as an extension, but we ignore this for now and call it deterministic. AD: yes
Unlike autoregressive models, \textsc{ChiroDiff} offers a way to draw conditional samples from the model without an explicit encoder when conditioned on homogeneous data (see section~\ref{sec:impl_conditioning}).
Yet another similar but important application we consider is \emph{stochastic vectorization}, i.e. sampling probable topological reconstruction $X$ given a perceptive input $\mathcal{R}(X)$ where $\mathcal{R}$ is a converter from vector representation to perceptive representation (e.g. raster image or point-cloud). We also learn deterministic mapping from noise to data with a variant of DDPM namely \emph{Denoising Diffusion Implicit Model} or DDIM which allows latent space interpolations like \citet{ha2017neural} and \citet{das2022sketchode}. A peculiar property of \textsc{ChiroDiff} allows a variant of the traditional interpolation which we term as ``Creative Mixing'', which do not require the model to be trained only on one end-points of the interpolation. We also show a number of unique use-cases like denoising/healing \citep{sketchhealing,shitong_denoising_score} and controlled abstraction \citep{umar_abstraction,das2022sketchode} in the context of chirographic data. As a drawback however, we loose some of the abilities of autoregressive models like stochastic completion etc.

In summary, we propose a Diffusion Model based framework, \textsc{ChiroDiff}, specifically suited for modelling continuous-time chirographic data (section~\ref{sec:creation}) which, so far, has predominantly been treated with autoregressive densities. Being non-autoregressive, \textsc{ChiroDiff} is capable of capturing holistic temporal concepts, leading to better reconstruction and generation metrics (section~\ref{sec:quant}). To this end, we propose the first diffusion model capable of handling temporally continuous data modality with variable length. We show a plethora of interesting and important downstream applications for chirographic data supported by \textsc{ChiroDiff} (section~\ref{sec:applications}).
% Specifically, we show tasks like stochastic vectorization (section~\ref{sec:stoch_vect}), implicit homogeneous conditioning (section~\ref{sec:impl_conditioning}), healing (section~\ref{sec:healing}), interpolation/mixing (section~\ref{sec:interp_mixing}) and controlled abstraction (section~\ref{sec:contr_abstraction}).

\section{Related Work}
% \vspace{-0.3cm}

Causal auto-regressive recurrent networks \citep{LSTM,gru} were considered to be a natural choice for sequential data modalities due to their inherent ability to encode ordering. It was the dominant tool for modelling natural language (NLP) \citep{bowman2016generating}, video \citep{srivastava2015unsupervised} and audio \citep{wavenet}. Recently, due to the breakthroughs in NLP \citep{transformer}, interests have shifted towards non-autoregressive models even for other modalities \citep{video_act_transformer,music_transformer}. Continuous-time chirographic models also experienced a similar shift in model class from LSTMs \citep{graves2013generating} to Transformers \citep{sketchformer,aksan2020cose} in terms of representation learning. Most of them however, still contains autoregressive generative components (e.g. causal transformers). Lately, \emph{set} structures have also been experimented with \citep{carlier2020deepsvg} for representing chirographic data as a collection of strokes. Due to difficulty with generating sets \citep{deepsets}, their objective function requires explicit accounting for mis-alignments. \textsc{ChiroDiff} finds a middle ground with the generative component being non-autoregressive while retaining the notion of order. \modified{A recent unpublished work \citep{diffusion_handwriting} applied diffusion models out-of-the-box to handwriting generation, although it lacks right design choices, explanations and extensive experimentation.}

Diffusion Models (DM), although existed for a while \citep{diffusionmodel}, made a breakthrough recently in generative modelling \citep{diffusionmodel_ho,diff_beat_gan,dalle_openai,glide_openai}. They are arguably by now the de-facto model for broad class of image generation \citep{diff_beat_gan} due to their ability to achieve both fidelity and diversity. With consistent improvements like efficient samplers \citep{ddim,plms_sampler}, latent-space diffusion \citep{latent_diffusion}, classifier(-free) guidance \citep{classifier_free_guidance,diff_beat_gan} these models are gaining traction in diverse set of vision-language (VL) problem. Even though DMs are generic in terms of theoretical formulation, very little focus have been given so far in non-image modalities \citep{audio_gen_diffusion,hoogeboom_mole_diff,geodiff}.

\section{Denoising Diffusion Probabilistic Models (DDPM)} \label{sec:diffusion}

DDPMs \citep{diffusionmodel_ho,diffusionmodel} are parametric densities realized by a stochastic ``reverse diffusion'' process that transforms a predefined isotropic gaussian prior $p(X_T) = \mathcal{N}(X_T; \mathbf{0}, \mathbf{I})$ into model distribution $p_{\theta}(X_0)$ by de-noising it in $T$ discrete steps. The sequence of $T$ parametric de-noising distributions admit \emph{markov property}, i.e. $p_{\theta}(X_{t-1} \vert X_{t:T}) = p_{\theta}(X_{t-1} \vert X_t)$ and can be chosen as gaussians \citep{diffusionmodel} as long as $T$ is large enough. With the model parameters defined as $\theta$, the de-noising conditionals have the form

\begin{equation} \label{eq:reverse_process}
    p_{\theta}(X_{t-1} \vert X_t) := \mathcal{N}(X_{t-1}; \bm{\mu}_{\theta}(X_t, t), \bm{\Sigma}_{\theta}(X_t, t))
\end{equation}

% The parametric model distribution $p_{\theta}(X_0)$ can be obtained by marginalizing the joint over all latent variables $X_{1:T}$, while the joint can be factored as $p_{\theta}(X_{0:T}) = p(X_T)\prod_{t=1}^T p_{\theta}(X_{t-1} \vert X_t)$.
Sampling can be performed with a DDPM sampler \citep{diffusionmodel_ho} by starting at the prior $X_T \sim p(X_T)$ and running ancestral sampling till $t = 0$ using $p_{\theta^*}(X_{t-1} \vert X_t)$ where $\theta^*$ denotes a set of trained model parameters.

% It is however, difficult to minimize the model negative log-likelihood expected over the data distribution $q(X_0)$, i.e. $\mathbb{E}_{q(X_0)} \left[ - \log p_{\theta}(X_0) \right]$ due to the presence of latent variables $X_{1:T}$. In order to mitigate this, Diffusion Models \emph{simulate} the latent variables by running a pre-defined ``forward diffusion'' process for $t = 0 \rightarrow T$, starting from $X_0 \sim q(X_0)$ and following a \emph{fixed} transition kernel

% \begin{equation*}
%     q(X_t \vert X_{t-1}) := \mathcal{N}(X_t; \sqrt{1-\beta_t} X_{t-1}; \beta_t \mathbf{I}).
% \end{equation*}

% For an appropriate choice of diffusion schedule $\beta_t$ over time $t$, this process guarantees $q(X_T)$ to match the model prior $p(X_T)$. With the simulated latent trajectory $X_{1:T} \sim q(X_{1:T} \vert X_0) = \prod_{t=1}^T q(X_t \vert X_{t-1})$, \citet{diffusionmodel} proved a variational bound (VB) on the expected model negative log-likelihood as $\mathcal{L}(\theta)$ which is easier to minimize

% \begin{equation} \label{eq:obj_simple_bound}
%     \mathcal{L}(\theta) \leq \mathcal{L}_{\mathrm{vb}}(\theta) = \mathbb{E}_{X_0 \sim q(X_0),\ X_{1:T} \sim q(X_{1:T} \vert X_0)} \big[ \log q(X_{1:T}\vert X_0) - \log p_{\theta}(X_{0:T}) \big]
% \end{equation}

Due to the presence of latent variables $X_{1:T}$, it is difficult to directly optimize the log-likelihood of the model $p_{\theta}(X_0)$. Practical training is done by first simulating the latent variables by a given ``forward diffusion'' process which allows sampling $X_{1:T}$ by means of

% due to the nature of the forward transition kernel $q(X_t \vert X_{t-1})$, the forward process allows sampling from $X_t$ without reliance on any other $t$ \citep{diffusionmodel_ho}, i.e. we can sample $X_t \sim q(X_t \vert X_0)$ where

\begin{equation} \label{eq:forward_reparam}
    q(X_t \vert X_0) = \mathcal{N}\left(X_t; \sqrt{\alpha_t} X_0; (1-\alpha_t )\mathbf{I}\right)
\end{equation}

with $\alpha_t \in (0, 1)$, a monotonically decreasing function of $t$, that completely specifies the forward noising process. By virtue of Eq.~\ref{eq:forward_reparam} and simplifying the reverse conditionals in Eq.~\ref{eq:reverse_process} with $\bm{\Sigma}_{\theta}(X_t, t) := \sigma_t^2 \mathbf{I}$, \citet{diffusionmodel,diffusionmodel_ho} derived an approximate variational bound $\mathcal{L}_{\mathrm{simple}}(\theta)$ that works well in practice

\begin{equation*}
    \mathcal{L}_{\mathrm{simple}}(\theta) = \mathbb{E}_{X_0\sim q(X_0),t\sim\mathcal{U}[1, T],\ \bm{\epsilon}\sim \mathcal{N}(\mathbf{0}, \mathbf{I})} \Big[ \vert\vert \bm{\epsilon} - \bm{\epsilon}_{\theta}(X_t(X_0, \bm{\epsilon}), t) \vert\vert^2 \Big]
\end{equation*}

where a reparameterized Eq.~\ref{eq:forward_reparam} is used to compute a ``noisy'' version of $X_0$ as $X_t(X_0, \bm{\epsilon}) = \sqrt{\alpha_t} X_0 + \sqrt{1-\alpha_t}\bm{\epsilon}$. Also note that the original parameterization of $\bm{\mu}_{\theta}(X_t, t)$ is modified in favour of $\bm{\epsilon}_{\theta}(X_t, t)$, an estimator to predict the noise $\bm{\epsilon}$ given a noisy sample $X_t$ at any step $t$. Please note that they are related as $\bm{\mu}_{\theta}(X_t, t) = \frac{1}{\sqrt{1-\beta_t}} \left( X_t - \frac{\beta_t}{\sqrt{1-\alpha_t}} \bm{\epsilon}_{\theta}(X_t, t) \right)$ where $\beta_t \triangleq 1 - \frac{\alpha_t}{\alpha_{t-1}}$.

% (see \citet{diffusionmodel_ho} for details). With trained model parameters $\theta^*$, samples can be drawn from the model as $X_{t-1} = \bm{\mu}_{\theta^*}(X_t, t) + \sigma_t^2 \bm{\epsilon}$ for $t = T \rightarrow 0$.

% \vspace{-0.6cm}
\section{Diffusion Model for chirographic data} \label{sec:creation}
% \vspace{-0.5cm}
Just like traditional approaches, we use the polyline sequence $X = \left[\ \cdots, \left(\mathbf{x}^{(j)}, p^{(j)} \right), \cdots\ \right]$ where the $j$-th point is $\mathbf{x}^{(j)} \in \mathbb{R}^2$ and $p^{(j)} \in \{ -1, 1 \}$ is a binary bit denoting the pen state, signaling an end of stroke. This representation is popularized by \citet{ha2017neural} and known as \emph{Three-point format}. We employ the same pre-processing steps (equispaced resampling, spatial scaling etc) laid down by \citet{ha2017neural}. Note that the cardinality of the sequence $|X|$ may vary for different samples.

\textsc{ChiroDiff} is fairly similar to the standard DDPM described in section~\ref{sec:diffusion}, with the sequence $X$ treated as a vector arranged by a particular topology. However, we found it beneficial not to directly use absolute points sequence $X$ but instead use velocities $V = \left[\ \cdots, \left(\mathbf{v}^{(j)}, p^{(j)} \right), \cdots\ \right]$, where $\mathbf{v}^{(j)} = \mathbf{x}^{(j+1)} - \mathbf{x}^{(j)}$ which can be readily computed using crude forward/backward differences. Upon generation, we can restore its original form by computing $\mathbf{x}^{(j)} = \sum_{j' \leq j} \mathbf{v}^{(j')}$. By modelling higher-order derivatives (velocity instead of position), the model focuses on high-level concepts rather than local temporal details \citep{ha2017neural,das2022sketchode}. We may use $X$ and $V$ interchangeably as they can be cheaply converted back and forth at any time.

Please note that we will use the subscript $t$ to denote the diffusion step and the superscript $^{(j)}$ to denote elements in the sequence. Following section~\ref{sec:diffusion}, we define \textsc{ChiroDiff}, our primary chirographic generative model $p_{\theta}(V)$ also as DDPM. We use a forward diffusion process termed as ``sequence-diffusion'' that diffuses each element $(\mathbf{v}^{(j)}_0, p^{(j)}_0)$ independently analogous to Eq.~\ref{eq:forward_reparam}

\begin{align*}
    q(V_t \vert V_0) = \textstyle{\prod_{j=1}^{|V|}} q(\mathbf{v}_t^{(j)}\vert& \mathbf{v}_0^{(j)}) \textstyle{\prod_{j=1}^{|V|}} q(p_t^{(j)}\vert p_0^{(j)}), \text{ with } \\
    q(\mathbf{v}_t^{(j)}\vert \mathbf{v}_0^{(j)}) = \mathcal{N}(\mathbf{v}^{(j)}_t; \sqrt{\alpha_t} \mathbf{v}^{(j)}_0, (1- \alpha_t) \mathbf{I})&,\ q(p_t^{(j)}\vert p_0^{(j)}) = \mathcal{N}(p^{(j)}_t; \sqrt{\alpha_t} p^{(j)}_0, (1 - \alpha_t) \mathbf{I})
\end{align*}

\begin{figure}[t]
    \centering
    \includegraphics[width=\linewidth]{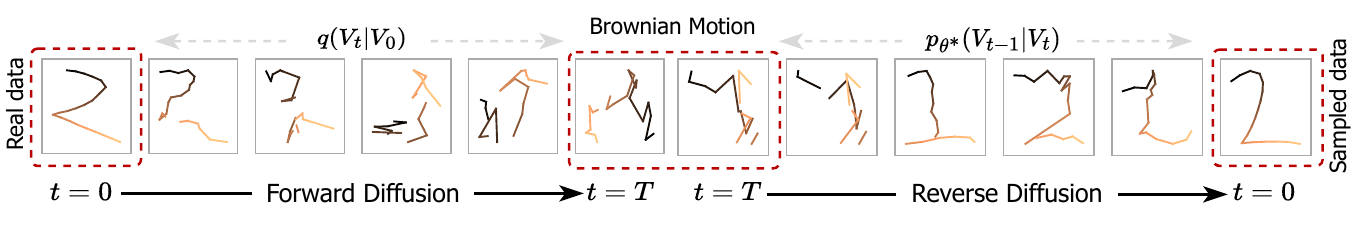}%
    \vspace{-0.3cm}
    \caption{The forward and reverse diffusion on chirographic data. The ``disconnected lines'' effect is due to the pen-bits being diffused together. We show the topology by color map (black to yellow).}
    % \vspace{-0.7cm}
    \label{fig:chiro_diff_viz}
\end{figure}

Consequently, the prior at $t=T$ has the form $q(V_T) = \prod_{j=1}^{|V|} q(\mathbf{v}_T^{(j)}) \prod_{j=1}^{|V|} q(p_T^{(j)})$ where individual elements are standard normal, i.e. $q(\mathbf{v}_T^{(j)}) = q(p_T^{(j)}) = \mathcal{N}(\mathbf{0}, \mathbf{I})$.
Note that we treat the binary pen state just like a continuous variable, an approach recently termed as ``analog bits'' by \citet{analog_bits_hinton}. Our experimentation show that it works quite well in practice. While generating, we map the analog bit to its original discrete states $\{-1, 1\}$ by simple thresholding at $p=0$.

With the \emph{reverse sequence diffusion} process modelled as parametric conditional gaussian kernels similar to section~\ref{sec:diffusion}, i.e. $\displaystyle{p_{\theta}(V_{t-1} \vert V_t) := \mathcal{N}(V_{t-1}; \bm{\mu}_{\theta}(V_t, t), \sigma_t^2 \mathbf{I})}$ and analogous change in parameterization (from $\bm{\mu}_{\theta}$ to $\bm{\epsilon}_{\theta}$), we can minimize the following loss

\begin{equation}\label{eq:loss_creation}
    \mathcal{L}_{\mathrm{simple}}(\theta) = \mathbb{E}_{V_0\sim q(V_0),\ t\sim\mathcal{U}[1, T],\ \bm{\epsilon}\sim \mathcal{N}(\mathbf{0}, \mathbf{I})} \Big[ \vert\vert \bm{\epsilon} - \bm{\epsilon}_{\theta}(V_t(V_0, \bm{\epsilon}), t) \vert\vert^2 \Big]
\end{equation}

With a trained $\bm{\epsilon}_{\theta^*}$, we can run DDPM sampler as $V_{t-1} \sim p_{\theta^*}(V_{t-1} \vert V_t)$ (refer to section~\ref{sec:diffusion}) iteratively for $t=T \rightarrow 1$. A deterministic variant, namely DDIM \citep{ddim}, can also be used as

\begin{equation}\label{eq:ddim_sampling}
    \displaystyle{
    V_{t-1} = \sqrt{\alpha_{t-1}} \left( \frac{V_t - \sqrt{1-\alpha_t} \bm{\epsilon}_{\theta^*}(V_t, t)}{\sqrt{\alpha_t}} \right) + \sqrt{1 - \alpha_{t-1}} \bm{\epsilon}_{\theta^*}(V_t, t)
    }
\end{equation}

\begin{figure}[t]
    \centering
    \includegraphics[width=\linewidth]{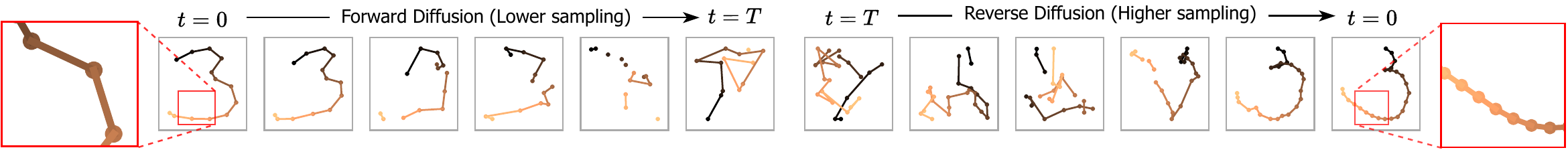}
    % \vspace{-0.5cm}
    \caption{The reverse process started with higher cardinality $|X|$ than the forward process.}
    % \vspace{-1.cm}
    \label{fig:arbitrary_sampling}
\end{figure}

Unlike the usual choice of U-Net in pixel-based perception models \citep{diffusionmodel_ho,diffusion_SDE_yangsong,dalle_openai,glide_openai}, \textsc{ChiroDiff} requires a sequence encoder as $\bm{\epsilon}_{\theta}(V_t, t)$ in order to preserve and utilize the ordering of elements. We chose to encode each element in the sequence with the entire sequence as context, i.e. $\bm{\epsilon}_{\theta}(\mathbf{v}^{(j)}_t, V_t, t)$. Two prominent choices for such functional form are Bi-directional RNN (Bi-RNN) and Transformer encoder \citep{settransformer} with positional embedding. We noticed that Bi-RNN works quite well and provides much faster and better convergence. A design choice we found beneficial is to concatenate the absolute positions $X_t$ along with $V_t$ to the model, i.e. $\bm{\epsilon}_{\theta}(\cdot, [V_t; X_t], t)$, exposing the model to the absolute state of the noisy data at $t$ instead of drawing dynamics only. Since $X_t$ can be computed from $V_t$ itself, we drop $X_t$ from the function arguments now onward just for notation brevity. Please note that the generation process is non-causal as it has access to the entire sequence while diffusing. This gives rise to a non-autoregressive model and thereby focusing on holistic concepts instead of low-level motor program. This design allows the reverse diffusion (generation) process to correct \emph{any} part of sequence from earlier mistakes, which is a not possible in auto-regressive models.

\keypoint{Transforming ``Brownian motion'' into ``Guided motion'':} \textsc{ChiroDiff}'s generation process has an interesting interpretation. Recall that the reverse diffusion process begins at $V_T = [\cdots, (\mathbf{v}_T^{(j)}, p_T^{(j)}), \cdots ]$ where each velocity element $\mathbf{v}_T^{(j)} \sim \mathcal{N}(\mathbf{0}, \mathbf{I})$. Due to our velocity-position encoding described above, the original chirographic structure is then $\mathbf{x}_T^{(j)} = \sum_{j'} \mathbf{v}_T^{(j')}$ which, by definition, is a discretized \emph{brownian motion} with unit step size. With the reverse process unrolled in time, the brownian motion with full randomness transforms into a motion with structure, leading to realistic data samples. We illustrate the entire process in Figure~\ref{fig:chiro_diff_viz}.

\keypoint{Length conditioned re-sampling:} A noticeable property of \textsc{ChiroDiff}'s generative process is that there is no \emph{hard} conditioning on the cardinality of the sequence $|X|$ or $|V|$ due to our choice of the parametric model $\bm{\epsilon}_{\theta}(\cdot, t)$. As a result, we can kick-off the generation (reverse diffusion) process by sampling from a prior $p(V_T) = \prod_{j=1}^{L} q(\mathbf{v}_T^{(j)}) q(p_T^{(j)})$ of any length $L$, potentially higher than what the model was trained on. We hypothesize and empirically show (in section~\ref{sec:quant}) that if trained to optimiality, the model indeed captures high level geometric concepts and can generate similar data with higher sampling rate (refer to Figure~\ref{fig:arbitrary_sampling}) with relatively less error. We credit this behaviour to the accessibility of the entire sequence $V_t$ (and additionally $X_t$) to the model $\bm{\epsilon}_{\theta}(\cdot)$. With the full sequence visible, the model can potentially build an internal (implicit) global representation which explains the resilience on increased temporal sampling resolution.

\section{Experiments \& Results}

\subsection{Datasets}
\vspace{-0.2cm}
\keypoint{VectorMNIST or VMNIST \citep{das2022sketchode}} is a vector analog of traditional MNIST digits dataset. It contains 10K samples of 10 digits ('0' to '9') represented in polyline sequence format. We use 80-10-10 splits for our all our experimentation.

\keypoint{KanjiVG\footnote{Original KanjiVG: \href{https://kanjivg.tagaini.net/}{kanjivg.tagaini.net}}} is a vector dataset containing Kanji characters. We use a preprocessed version of the dataset\footnote{Pre-processed KanjiVG: \href{https://github.com/hardmaru/sketch-rnn-datasets/tree/master/kanji}{github.com/hardmaru/sketch-rnn-datasets/tree/master/kanji}} which converted the original SVGs into polyline sequences. This dataset is used in order to evaluate our method's effective on complex chirographic structures with higher number of strokes.

\keypoint{\emph{Quick, Draw!} \citep{ha2017neural}} is the largest collection of free-hand doodling dataset with casual depictions of given concepts. This dataset is an ideal choice for evaluating a method's effectiveness on real noisy data since it was collected by means of large scale crowd-sourcing. In this paper, we use the following categories: \{cat, crab, bus, mosquito, fish, yoga, flower\}.

\subsection{Implementation Details}

\textsc{ChiroDiff}'s forward process, just like traditional DDPMs, uses a linear noising schedule of $\left\{\beta_{\mathrm{min}} = 10^{-4} \cdot 1000/T, \beta_{\mathrm{max}} = 2\times 10^{-2} \cdot 1000/T \right\}$ as found out by \citet{iddpm,diff_beat_gan} to be quite robust. We noticed that there isn't much performance difference with different diffusion lengths, so we choose a standard value of $T=1000$. The parametric noise estimator $\bm{\epsilon}_{\theta}(\mathbf{v}_t^{(j)}, V_t, t)$ is chosen to be a bi-directional GRU encoder \citep{gru} where each element of the sequence is encoded while having contextual information from both direction of the sequence, making it non-causal. We use a 2-layer GRU with $D=48$ hidden units for VMNIST and 3-layer GRU for QuickDraw ($D=128$) and KanjiVG ($D=96)$. We also experimented with transformers with positional encoding but failed to achieve reasonable results, concluding that positional encoding is not a good choice for representing continuous time. We trained all of our models by minimizing Eq.~\ref{eq:loss_creation} using AdamW optimizer \citep{adamw} and step-wise LR scheduling of $\gamma_e = 0.9997 \cdot \gamma_{e-1}$ at every epoch $e$ where $\gamma_0 = 6\times 10^{-3}$. The diffusion time-step $t \in {1, 2, \cdots, T}$ was made available to the model by concatenating sinusoidal positional embeddings \citep{transformer} into each element of a sequence at every layer. We noticed the importance of reverse process variance $\sigma_t^2$ in terms of generation quality of our models. We found $\sigma_t^2 = 0.8 \Tilde{\beta}_t$ to work well in majority of the cases, where $\Tilde{\beta}_t = \frac{1-\alpha_{t-1}}{1-\alpha_t} \beta_t$ as defined by \citet{diffusionmodel_ho} to be true variance of the forward process posterior. \modified{Please refer to the project page\footnote{Our project page: \href{https://ayandas.me/chirodiff}{https://ayandas.me/chirodiff}} for full source code.}
% We implemented the framework in PyTorch Lightning \citep{Falcon_PyTorch_Lightning_2019} and ran our experiments on several RTX3090Ti/RTX2080Ti GPUs.

\subsection{Quantitative evaluations} \label{sec:quant}

In order to assess the effectiveness of our model for chirographic data, we perform quantitative evaluations and compare with relevant approaches. We measure performance in terms of representation learning, generative modelling and computational efficiency. By choosing proper dimensions for competing methods/architectures, we ensured approximately same model capacity (\# of parameters) for fare comparison.

\begin{figure}
    \centering
    \includegraphics[width=0.97\linewidth]{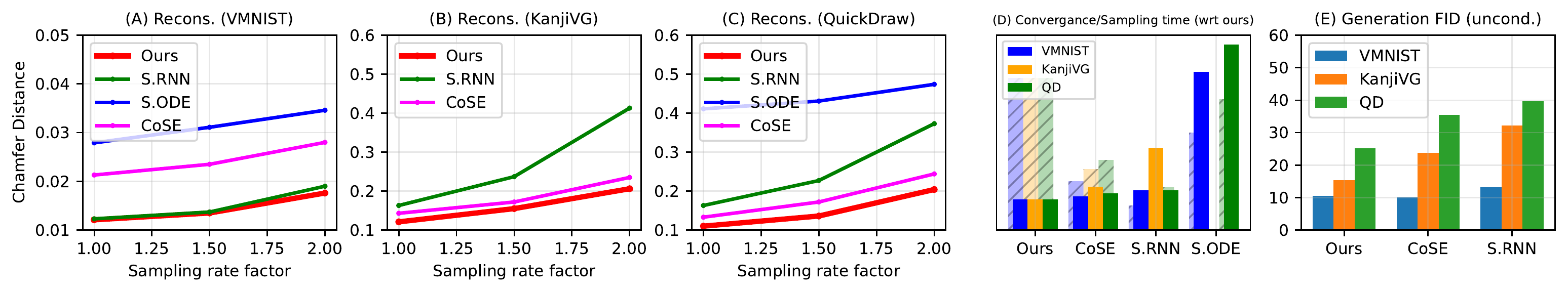}
    \vspace{-0.5cm}
    \caption{(A, B, C) Reconstruction CD against sampling rate factor. (D) Relative convergence time \& sampling time (transparent bars) w.r.t our method. (E) FID of unconditional generation (averaged over multiple classes for QD).}
    \label{fig:quant}
    \vspace{-0.5cm}
\end{figure}

\begin{wrapfigure}[14]{R}{0.6\textwidth}
    \includegraphics[width=\linewidth]{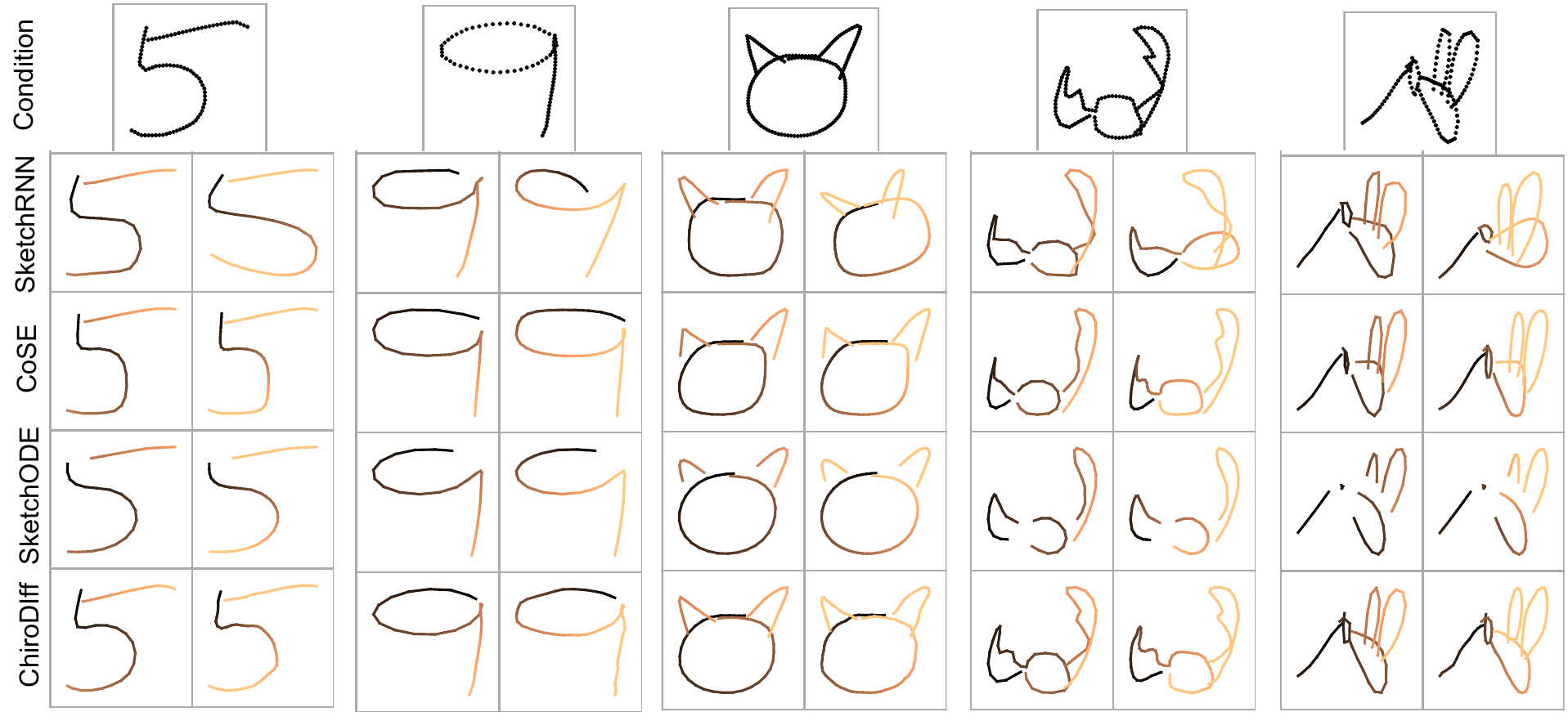}
    \vspace{-0.5cm}
    \caption{$1^{st}$ and $2^{nd}$ column for each example depicts sampling rate $1$ \& $2$ respectively while reconstructing.}
    \label{fig:qual_recon}
\end{wrapfigure}

\keypoint{Reconstruction} We construct a conditional variant of \textsc{ChiroDiff} with an encoder $\mathcal{E}_V$ being a traditional Bi-GRU for fully encoding a given data sample $V$ into latent code $\mathbf{z}$. The decoder, in our case, is a diffusion model described in section~\ref{sec:creation}. We sample from the conditional model $p_{\theta}(V_0\vert \mathbf{z} = \mathcal{E}_V(V))$ which is effectively same as standard DDPM but with the noise-estimator $\bm{\epsilon}_{\theta^*}(V_t, t, \mathbf{z})$ additionally conditioned on $\mathbf{z}$. We expose the latent variable $\mathbf{z}$ to the noise-estimator by simply concatenating it with every element $j$ at all timestep $t \in [1, T]$. We also evaluate \textsc{ChiroDiff}'s ability to adopt to higher temporal resolution while sampling, proving our hypothesis that it captures concepts at a holistic level. We encode a sample with $\mathcal{E}_V$, and decode explicitly with a higher temporal sampling rate (refer to section~\ref{sec:creation}). We compare our method with relevant frameworks like SketchODE \citep{das2022sketchode}, SketchRNN \citep{ha2017neural} and CoSE \citep{aksan2020cose}. Since autoregressive models like SketchRNN has no explicit way to increase temporal resolution, we train different models with resampled data, which is already disadvantageous. We quantitatively compare them with \emph{Chamfer Distance (CD)} \citep{pointnet} (ignore the pen-up bit) for conditional reconstruction. Figure~\ref{fig:quant} shows the reconstruction CD against sampling rate factor (multiple of the original data cardinality) which shows the resilience of our model against higher sampling rate. SketchRNN being autoregressive, fails at high sampling rate (longer sequences), as visible in Figure~\ref{fig:quant}(A, B \& C). CoSE and SketchODE being naturally continuous, has a relatively flat curve. Also, we couldn't reasonably train SketchODE on the complex KanjiVG dataset (due to computational and convergance issues) and hence omitted from Figure~\ref{fig:quant}. Qualitative examples of reconstruction shown in Fig.~\ref{fig:qual_recon}.

\keypoint{Generation} We assess the generative performance of \textsc{ChiroDiff} by sampling unconditionally (in $50$ steps with DDIM sampler) and compute the FID score against the real data samples. Since the original inception network is not trained on chirographic data, we train our own on \emph{Quick, Draw!} dataset \citep{ha2017neural} following the setup of \citep{creativesketch_doodlergan}. We compare our method with SketchRNN \citep{ha2017neural} and CoSE \citep{aksan2020cose} on all three datasets. Quantitative results in Figure~\ref{fig:quant}(E) show a consistent superiority of our model in terms of generation FID. QuickDraw FID values are averaged over the individual categories used. Qualitative samples are shown in Figure~\ref{fig:banner}.

\keypoint{Computational Efficiency} We also compare our method with competing methods in terms of ease of training and convergence. We found that our method, being from Diffusion Model family, enjoys easy training dynamics and relatively faster convergence (refer to Figure~\ref{fig:quant}(D)). We also provide approximate sampling time for unconditional generation.

\subsection{Downstream Applications} \label{sec:applications}

\subsubsection{Stochastic vectorization} \label{sec:stoch_vect}

\begin{wrapfigure}[14]{R}{0.33\textwidth}
    \vspace{-0.5cm}
    \includegraphics[width=\linewidth]{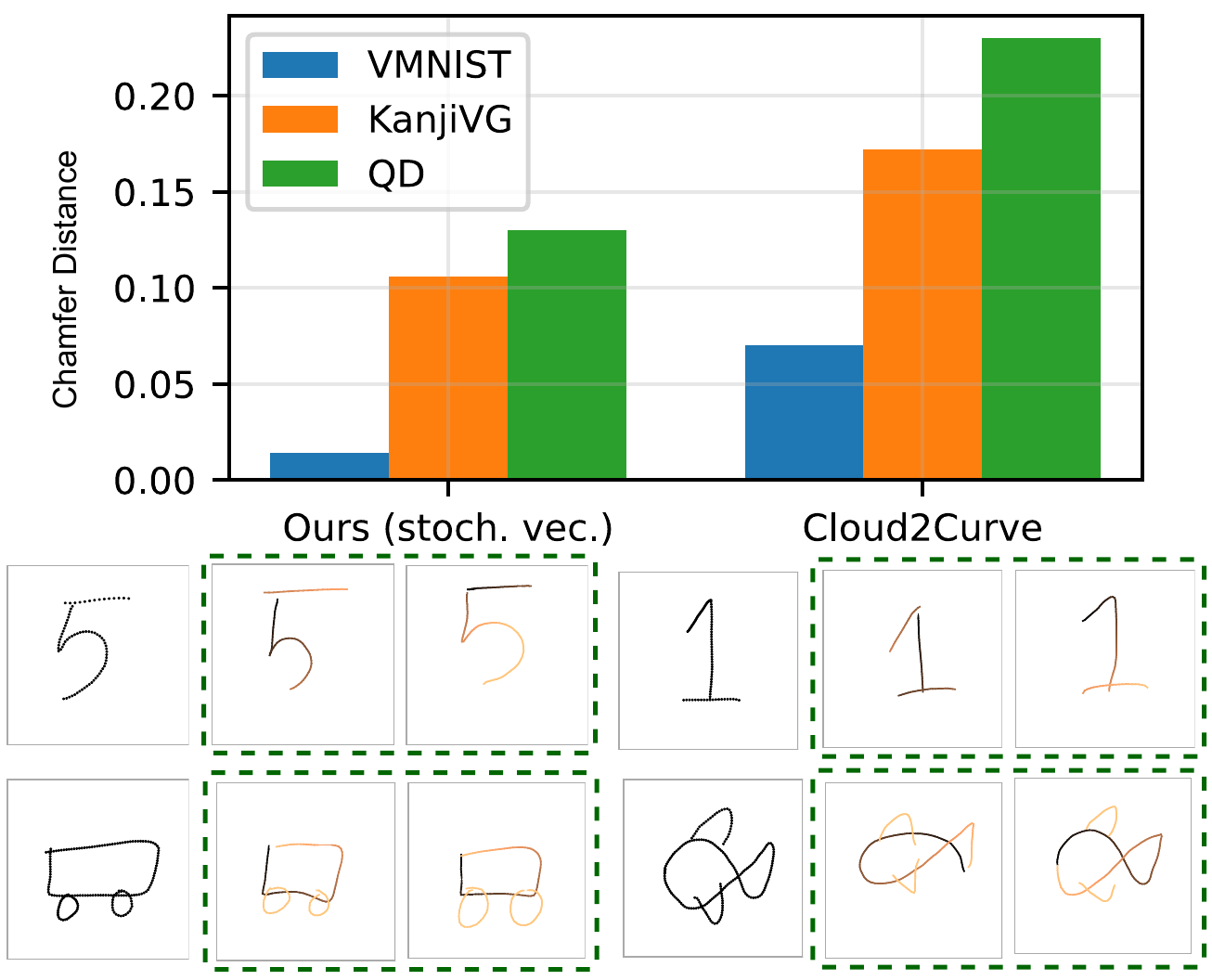}%
    \vspace{-0.3cm}
    \caption{Stochastic vectorization. Note the different topologies (color map) of the samples.}
    \vspace{-0.4cm}
    \label{fig:stoch_vec}
\end{wrapfigure}

An interesting use-case of generative chirographic model is \emph{stochastic vectorization}, i.e. recreating plausible chirographic structure (with topology) from a given perceptive input. This application is intriguing due to human's innate ability to do the same with ease. The recent success of diffusion models in capturing distributions with high number of modes \citep{dalle_openai,glide_openai} prompted us to use it for stochastic vectorization, a problem of inferring potentially multi-modal distribution. We simply condition our generative model on a perceptive input $\mathcal{X} = \mathcal{R}(X) = \{ \mathbf{x}^{(j)} \vert (\mathbf{x}^{(j)}, p^{(j)}) \in X \}$, i.e. we convert the sequence into a point-set (also densely resample them as part of pre-processing). We employ a set-transformer encoder $\mathcal{E}_R(\cdot)$ with max-pooling aggregator \citep{settransformer} to obtain a latent vector $\mathbf{z}$ and condition the generative model similar to section~\ref{sec:quant}, i.e. $p_{\theta}(V\vert \mathbf{z} = \mathcal{E}_R(\mathcal{X}))$. We evaluated the conditional generation with Chamfer Distance (CD) on test set and compare with \citet{cloud2curvedas} (refer to Figure~\ref{fig:stoch_vec}).

\subsubsection{Implicit Conditioning} \label{sec:impl_conditioning}

Unlike explicit conditioning mechanism (includes an encoder) described in section~\ref{sec:quant}, \textsc{ChiroDiff} allows a form of ``Implicit Conditioning'' which requires no explicit encoder. Such conditioning is more stochastic and may not be used for reconstruction, but can be used to sample similar data from a pre-trained model $p_{\theta^*}$. Given a condition $X_0^{\mathrm{cond}}$ (or $V^{\mathrm{cond}}_0$), we sample a noisy version at $t=T_c < T$ (a hyperparameter) using the forward diffusion (Eq.~\ref{eq:forward_reparam}) process as $V_{T_c}^{\mathrm{cond}} = \sqrt{\alpha_{T_c}} V_0^{\mathrm{cond}} + \sqrt{1 - \alpha_{T_c}} \bm{\epsilon}$ where $\bm{\epsilon} \sim \mathcal{N}(\mathbf{0}, \mathrm{I})$. We then utilize the trained model $p_{\theta^*}$ to gradually de-noise $V_{T_c}^{\mathrm{cond}}$. We run the reverse process from $t=T_c$ till $t=0$

\vspace{-.3cm}
\begin{equation*}
    V_{t-1} \sim p_{\theta^*}(V_{t-1}\vert V_t)\text{, for } T_c > t > 0\text{, with } V_{T_c} := V_{T_c}^{\mathrm{cond}}
\end{equation*}

The hyperparameter $T_c$ controls how much the generated samples correlate the given condition. By starting the reverse process at $t=T_c$ with the noisy condition, we essentially restrict the generation to be within a region of the data space that resembles the condition. Higher the value of $T_c$, the generated samples will resemble the condition more (refer to Figure~\ref{fig:impl_cond_healing}). We also classified the generated samples for VMNIST \& \emph{Quick, Draw!} and found it to belong to the same class as the condition $93\%$ of the time in average.

% We use the term ``Implicit Conditioning'' to refer to conditional generative models for which there is no explicit encoder for encoding the condition. Instead, the generation process for a pre-trained model is tweaked to satisfy the conditioning. Such implicit conditioning is more popular in Diffusion Models family. some tasks are just not possible in other model family.

\subsubsection{Healing} \label{sec:healing}

\begin{wrapfigure}[13]{R}{0.5\textwidth}
    \vspace{-0.5cm}
    \includegraphics[width=\linewidth]{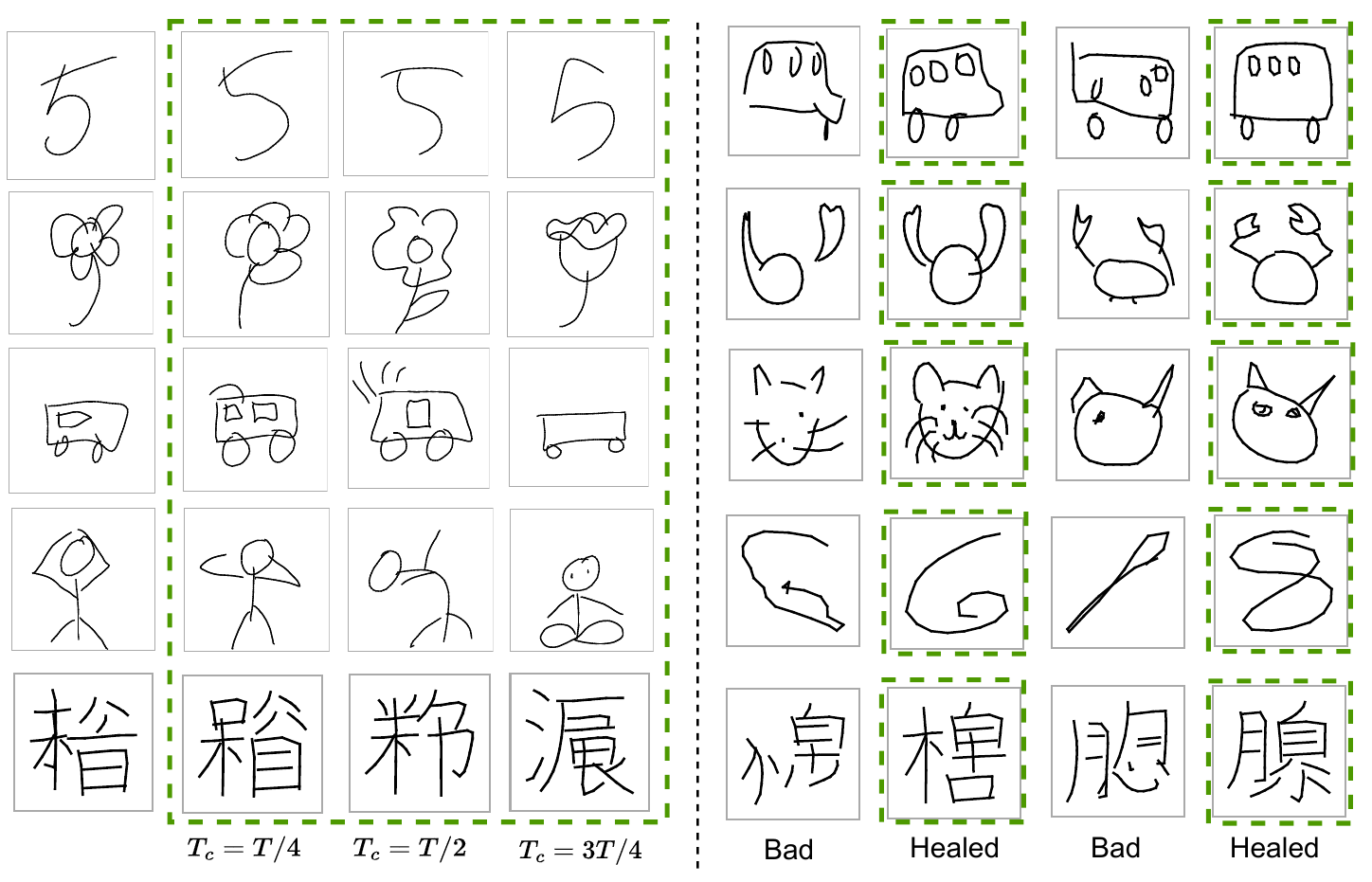}%
    \vspace{-0.3cm}
    \caption{Implicit conditioning and healing.}
    \label{fig:impl_cond_healing}
\end{wrapfigure}

The task of healing is more prominent in 3D point-cloud literature \citep{shitong_denoising_score}. Even though the typical chirographic (autoregressive) models offer ``stochastic completion'' \citep{ha2017neural}, it does not offer an easy way to ``heal'' a sample due to uni-directional generation. It is only very recently, works have emerged that propose tools for healing bad sketches \citep{sketchhealing}. With diffusion model family, it is fairly straightforward to solve this problem with \textsc{ChiroDiff}. Given a ``poor'' chirographic data $\Tilde{X}_0$, we would like to generate samples from a region in $p_{\theta^*}(X_0)$ close to $\Tilde{X}_0$ in terms of semantic concept.
Surprisingly, this problem can be solved with the ``Implicit Conditioning'' described in section~\ref{sec:impl_conditioning}. Instead of a real data as condition, we provide the poorly drawn data $\Tilde{X}_0$ (equivalently $\Tilde{V}_0$) as condition. Just as before, we run the reverse process starting at $t=T_h$ with $V_{T_h} := \Tilde{V}_{T_h}$ in order to sample from a healed data distribution around $\Tilde{V}_0$. $T_h$ is a similar hyperparameter that decides the trade-off between healing the given sample and drifting away from it in terms of high-level concept. Refer to Figure~\ref{fig:impl_cond_healing} (right) for qualitative samples of healing (with $T_h = T/5$).

\subsubsection{Creative Mixing} \label{sec:interp_mixing}

Creative Mixing is a chirographic task of merging two high-level concepts into one. This task is usually implemented as latent-space interpolation in traditional autoencoder-style generative models \citep{ha2017neural,das2022sketchode}. A variant of \textsc{ChiroDiff} that uses DDIM sampler \citep{ddim}, can be used for similar interpolations. We use a pre-trained conditional model to decode the interpolated latent vector using $V_T=\mathbf{0}$ as a fixed point. Given two samples $V_{0_1}$ and $V_{0_2}$, we compute the interpolated latent variable as $\mathbf{z}_{\mathrm{interp}} = (1 - \delta) \mathcal{E}_V(V_{0_1}) + \delta \mathcal{E}_V(V_{0_2})$ for any $\delta \in [0, 1]$ and run DDIM sampler shown in Eq.~\ref{eq:ddim_sampling} with the noise-estimator $\bm{\epsilon}_{\theta^*}(V_t, t, \mathbf{z}_{\mathrm{interp}})$.

% \begin{equation*} %\label{eq:ddim_sampling}
%     \displaystyle{
%     V_{t-1} = \sqrt{\alpha_{t-1}} \left( \frac{V_t - \sqrt{1-\alpha_t} \bm{\epsilon}_{\theta^*}(V_t, t, \mathbf{z}_{\mathrm{interp}})}{\sqrt{\alpha_t}} \right) + \sqrt{1 - \alpha_{t-1}} \bm{\epsilon}_{\theta^*}(V_t, t, \mathbf{z}_{\mathrm{interp}})
%     }
% \end{equation*}

\begin{figure}[t]
    \centering
    \includegraphics[width=0.98\linewidth]{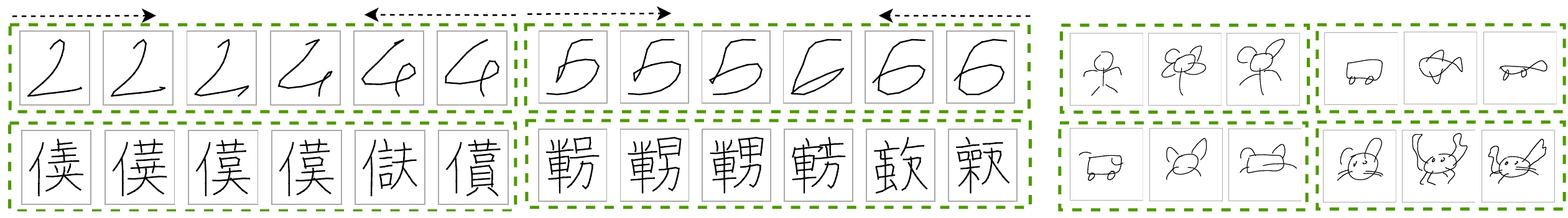}
    \vspace{-0.4cm}
     \caption{(Left) Latent-space semantic interpolation with DDIM. (Right) Creative Mixing shown as three-tuples consisting of $X_0$, $X_0^{\mathrm{ref}}$ and the mixed sample respectively.}
     \vspace{-0.3cm}
    \label{fig:mixing}
\end{figure}

This solution works well for some datasets (KanjiVG \& VMNIST; shown in Figure~\ref{fig:mixing} (left)). For others (\emph{Quick, Draw}), we instead propose a more general method inspired by ILVR \citep{ilvr_ddpm} that allows ``mixing'' using the DDPM sampler itself. In fact, it allows us to perform mixing \emph{without} one of the samples (reference sample) being known to the trained model. Given two samples of potentially different concepts, $X_0$ and $X^{\mathrm{ref}}_0$ (or equivalently $V_0$ and $V^{\mathrm{ref}}_0$), we sample from a pre-trained conditional model given $V_0$, but with a modified reverse process
\vspace{-0.1cm}
\begin{equation*}\begin{split}
    X_{t-1} &= X'_{t-1} - \Phi_{\omega}(X'_{t-1}) + \Phi_{\omega}(X_{t-1}^{\mathrm{ref}}) \\
    \text{where, } V'_{t-1} &\sim p_{\theta^*}(V_{t-1} \vert V_t, \mathbf{z} = \mathcal{E}_V(V_0)\text{, and } V_{t-1}^{\mathrm{ref}} \sim q(V_{t-1}^{\mathrm{ref}} \vert V_0^{\mathrm{ref}})
    \label{eq:ilvr_mixing}
    \vspace{-.1cm}
\end{split}\end{equation*}

where $\Phi_{\omega}(\cdot)$ is a \emph{temporal} low-pass filter that reduces high-frequency details of the input data along \emph{temporal axis}. We implement $\Phi_{\omega}(\cdot)$ using temporal 1D convolution of window size $\omega = 7$. Please note that for the operation in Eq.~\ref{eq:ilvr_mixing} to be valid, the sequences must be of same length. We simply resample the conditioning sequence to match the cardinality of the reverse process. The three-tuples presented in Figure~\ref{fig:mixing}(right) shows the mixing between different categories of \emph{Quick, Draw!}.

\begin{wrapfigure}[12]{L}{0.6\textwidth}
    \vspace{-0.5cm}
    \includegraphics[width=\linewidth]{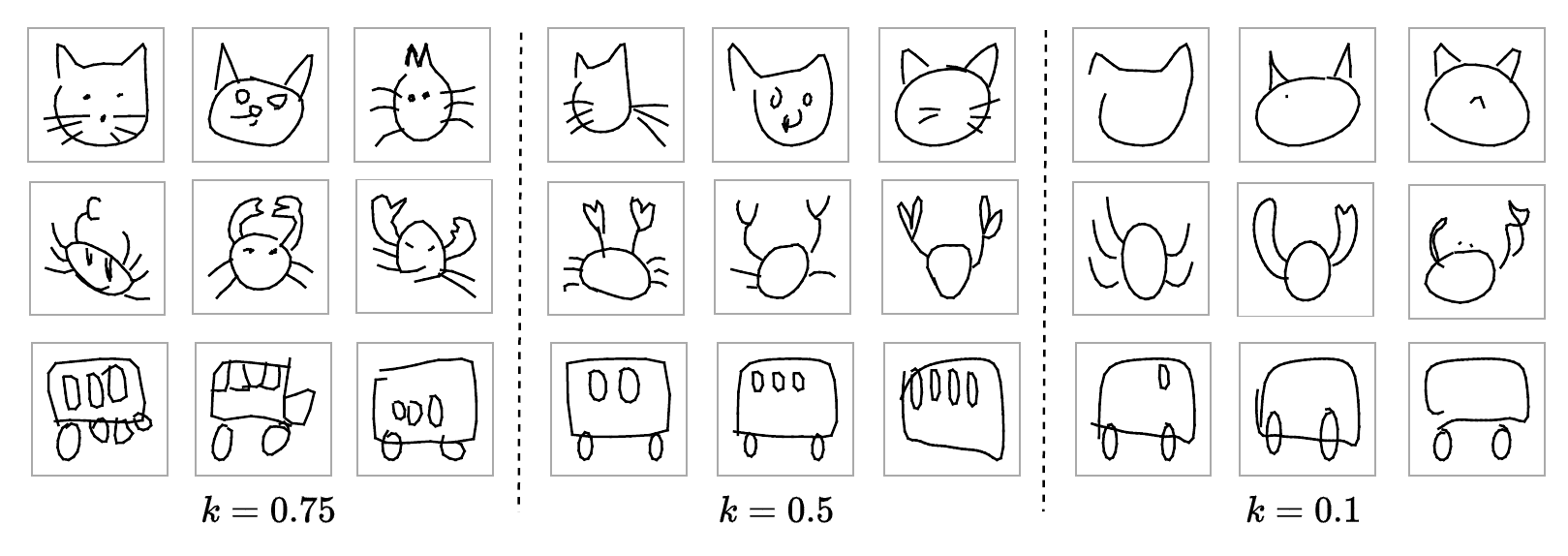}%
    \vspace{-0.3cm}
    \caption{With reduced reverse process variance $\sigma_t^2 = k\cdot \Tilde{\beta}$, generated samples loose high-frequency details but retains abstract concepts.}
    \label{fig:abstraction}
\end{wrapfigure}

\subsubsection{Controlled Abstraction} \label{sec:contr_abstraction}

\emph{Visual abstraction} is a relatively new task \citep{umar_abstraction,das2022sketchode} in chirographic literature that refers to deriving a new distribution (possibly with a control) that \emph{holistically} matches the data distribution, but more ``abstracted'' in terms of details. Our definition to the problem matches that of \citet{das2022sketchode}, but with an advantage of being able to use the same model instead of re-training for different controls.

The solution of the problem lies in the sampling process of \textsc{ChiroDiff}, which has an abstraction effect when the reverse process variance $\sigma_t^2$ is low. We define a continuous control $k \in [0, 1]$ as hyperparameter and use a reverse process variance of $\sigma_t^2 = k \cdot \Tilde{\beta}_t$. The rationale behind this method is that when $k$ is near zero, the reverse process stops exploring the data distribution and converges near the \emph{dominant modes}, which are data points conceptually resembling original data but more ``canonical'' representation (highly likely under data distribution). Figure~\ref{fig:abstraction} shows qualitative results of this observation on \emph{Quick, Draw!}.

\section{Conclusions, Limitations \& Future Work}

\begin{wrapfigure}[10]{R}{0.26\textwidth}
    \vspace{-0.5cm}
    \includegraphics[width=\linewidth]{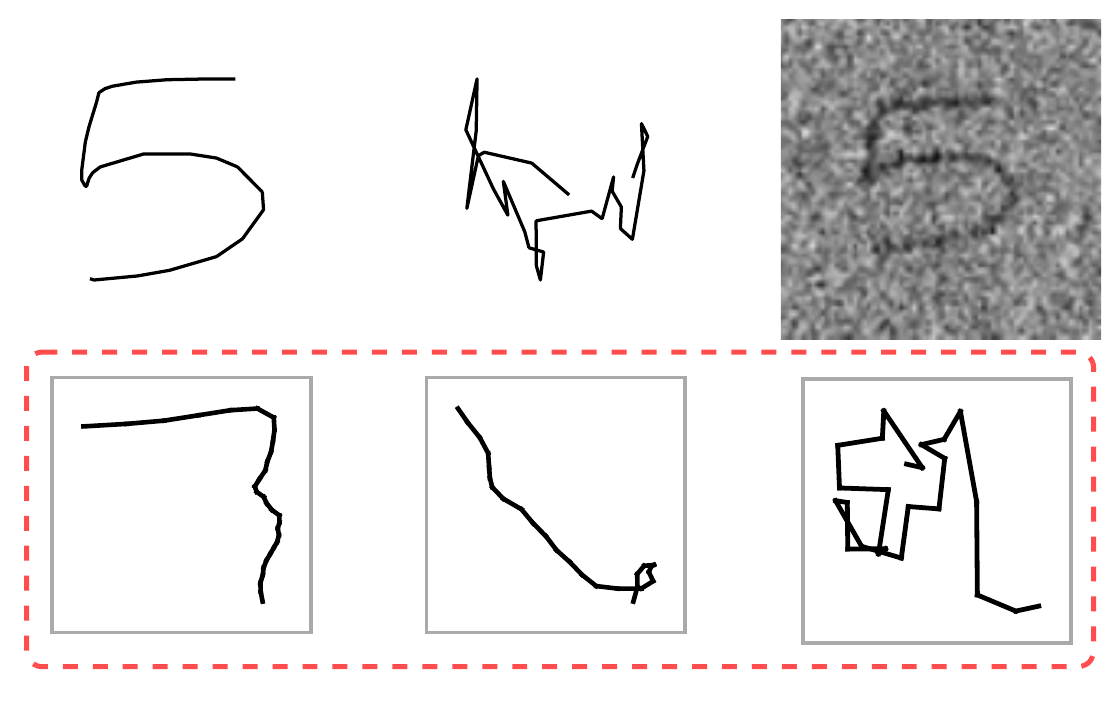}%
    \vspace{-0.3cm}
    \caption{\small{(Top) Noisy vector and raster data at same $\alpha$. (Bottom) Failures due to noise.}}
    \vspace{-0.3cm}
    \label{fig:limitations}
\end{wrapfigure}

\vspace{-.1cm}
In this paper, we introduced \textsc{ChiroDiff}, a non-autoregressive generative model for chirographic data. ChiroDiff is powered by DDPM and offers better holistic modelling of concepts that benefits many downstream tasks, some of which are not feasible with competing autoregressive models.
One limitation of our approach is that vector representations are more susceptible to noise than raster images (see Figure~\ref{fig:limitations} (Top)). Since we empirically set the reverse (generative) process variance $\sigma_t^2$, the noise sometimes overwhelms the model-predicted mean (see Figure~\ref{fig:limitations} (Bottom)). Moreover, due the use of velocities, the \emph{accumulated} absolute positions also accumulate the noise in proportion to its cardinality. One possible solution is to modify the noising process to be adaptive to the generation or the data cardinality itself. We leave this as a potential future improvement on \textsc{ChiroDiff}.

\newpage

\bibliography{main}
\bibliographystyle{iclr2023_conference}

\end{document}

%% file: math_commands.tex
\usepackage{amsmath,amsfonts,bm}

\def\eqref#1{equation~\ref{#1}}
\def\1{\bm{1}}

\DeclareMathAlphabet{\mathsfit}{\encodingdefault}{\sfdefault}{m}{sl}
\SetMathAlphabet{\mathsfit}{bold}{\encodingdefault}{\sfdefault}{bx}{n}

%% file: main.bbl
\begin{thebibliography}{53}
\providecommand{\natexlab}[1]{#1}
\providecommand{\url}[1]{\texttt{#1}}
\expandafter\ifx\csname urlstyle\endcsname\relax
  \providecommand{\doi}[1]{doi: #1}\else
  \providecommand{\doi}{doi: \begingroup \urlstyle{rm}\Url}\fi

\bibitem[Aksan et~al.(2020)Aksan, Deselaers, Tagliasacchi, and
  Hilliges]{aksan2020cose}
Emre Aksan, Thomas Deselaers, Andrea Tagliasacchi, and Otmar Hilliges.
\newblock Cose: Compositional stroke embeddings.
\newblock \emph{NeurIPS}, 2020.

\bibitem[Bowman et~al.(2016)Bowman, Vilnis, Vinyals, Dai, Jozefowicz, and
  Bengio]{bowman2016generating}
Samuel~R. Bowman, Luke Vilnis, Oriol Vinyals, Andrew Dai, Rafal Jozefowicz, and
  Samy Bengio.
\newblock Generating sentences from a continuous space.
\newblock In \emph{CoNLL}, 2016.

\bibitem[Cai et~al.(2020)Cai, Yang, Averbuch{-}Elor, Hao, Belongie, Snavely,
  and Hariharan]{shapegf}
Ruojin Cai, Guandao Yang, Hadar Averbuch{-}Elor, Zekun Hao, Serge~J. Belongie,
  Noah Snavely, and Bharath Hariharan.
\newblock Learning gradient fields for shape generation.
\newblock In \emph{ECCV}, 2020.

\bibitem[Carlier et~al.(2020)Carlier, Danelljan, Alahi, and
  Timofte]{carlier2020deepsvg}
Alexandre Carlier, Martin Danelljan, Alexandre Alahi, and Radu Timofte.
\newblock Deepsvg: A hierarchical generative network for vector graphics
  animation.
\newblock \emph{NeurIPS}, 2020.

\bibitem[Chen et~al.(2018)Chen, Rubanova, Bettencourt, and Duvenaud]{neuralode}
Tian~Qi Chen, Yulia Rubanova, Jesse Bettencourt, and David Duvenaud.
\newblock Neural ordinary differential equations.
\newblock In \emph{NeurIPS}, 2018.

\bibitem[Chen et~al.(2022)Chen, Zhang, and Hinton]{analog_bits_hinton}
Ting Chen, Ruixiang Zhang, and Geoffrey~E. Hinton.
\newblock Analog bits: Generating discrete data using diffusion models with
  self-conditioning.
\newblock \emph{ArXiv}, abs/2208.04202, 2022.

\bibitem[Cho et~al.(2014)Cho, van Merrienboer, G{\"{u}}l{\c{c}}ehre, Bahdanau,
  Bougares, Schwenk, and Bengio]{gru}
Kyunghyun Cho, Bart van Merrienboer, {\c{C}}aglar G{\"{u}}l{\c{c}}ehre, Dzmitry
  Bahdanau, Fethi Bougares, Holger Schwenk, and Yoshua Bengio.
\newblock Learning phrase representations using {RNN} encoder-decoder for
  statistical machine translation.
\newblock In \emph{EMNLP}, 2014.

\bibitem[Choi et~al.(2021)Choi, Kim, Jeong, Gwon, and Yoon]{ilvr_ddpm}
Jooyoung Choi, Sungwon Kim, Yonghyun Jeong, Youngjune Gwon, and Sungroh Yoon.
\newblock {ILVR:} conditioning method for denoising diffusion probabilistic
  models.
\newblock In \emph{ICCV}, 2021.

\bibitem[Das et~al.(2020)Das, Yang, Hospedales, Xiang, and
  Song]{das2020bziersketch}
Ayan Das, Yongxin Yang, Timothy Hospedales, Tao Xiang, and Yi-Zhe Song.
\newblock Béziersketch: A generative model for scalable vector sketches.
\newblock In \emph{ECCV}, 2020.

\bibitem[Das et~al.(2021)Das, Yang, Hospedales, Xiang, and
  Song]{cloud2curvedas}
Ayan Das, Yongxin Yang, Timothy~M. Hospedales, Tao Xiang, and Yi{-}Zhe Song.
\newblock Cloud2curve: Generation and vectorization of parametric sketches.
\newblock In \emph{CVPR}, 2021.

\bibitem[Das et~al.(2022)Das, Yang, Hospedales, Xiang, and
  Song]{das2022sketchode}
Ayan Das, Yongxin Yang, Timothy~M Hospedales, Tao Xiang, and Yi-Zhe Song.
\newblock Sketchode: Learning neural sketch representation in continuous time.
\newblock In \emph{ICLR}, 2022.

\bibitem[Dhariwal \& Nichol(2021)Dhariwal and Nichol]{diff_beat_gan}
Prafulla Dhariwal and Alexander~Quinn Nichol.
\newblock Diffusion models beat gans on image synthesis.
\newblock In \emph{NeurIPS}, 2021.

\bibitem[Ge et~al.(2021)Ge, Goswami, Zitnick, and
  Parikh]{creativesketch_doodlergan}
Songwei Ge, Vedanuj Goswami, Larry Zitnick, and Devi Parikh.
\newblock Creative sketch generation.
\newblock In \emph{ICLR}, 2021.

\bibitem[Gervais et~al.(2020)Gervais, Deselaers, Aksan, and
  Hilliges]{dididataset}
Philippe Gervais, Thomas Deselaers, Emre Aksan, and Otmar Hilliges.
\newblock The {DIDI} dataset: Digital ink diagram data.
\newblock \emph{CoRR}, 2020.

\bibitem[Girdhar et~al.(2019)Girdhar, Carreira, Doersch, and
  Zisserman]{video_act_transformer}
Rohit Girdhar, Jo{\~{a}}o Carreira, Carl Doersch, and Andrew Zisserman.
\newblock Video action transformer network.
\newblock In \emph{CVPR}, 2019.

\bibitem[Graves(2013)]{graves2013generating}
Alex Graves.
\newblock Generating sequences with recurrent neural networks.
\newblock \emph{arXiv preprint arXiv:1308.0850}, 2013.

\bibitem[Ha \& Eck(2018)Ha and Eck]{ha2017neural}
David Ha and Douglas Eck.
\newblock A neural representation of sketch drawings.
\newblock In \emph{ICLR}, 2018.

\bibitem[Ho \& Salimans(2022)Ho and Salimans]{classifier_free_guidance}
Jonathan Ho and Tim Salimans.
\newblock Classifier-free diffusion guidance.
\newblock \emph{CoRR}, 2022.

\bibitem[Ho et~al.(2020)Ho, Jain, and Abbeel]{diffusionmodel_ho}
Jonathan Ho, Ajay Jain, and Pieter Abbeel.
\newblock Denoising diffusion probabilistic models.
\newblock In \emph{NeurIPS}, 2020.

\bibitem[Hochreiter \& Schmidhuber(1997)Hochreiter and Schmidhuber]{LSTM}
Sepp Hochreiter and J{\"{u}}rgen Schmidhuber.
\newblock Long short-term memory.
\newblock \emph{Neural Comput.}, 1997.

\bibitem[Hoogeboom et~al.(2022)Hoogeboom, Satorras, Vignac, and
  Welling]{hoogeboom_mole_diff}
Emiel Hoogeboom, Victor~Garcia Satorras, Cl{\'{e}}ment Vignac, and Max Welling.
\newblock Equivariant diffusion for molecule generation in 3d.
\newblock In Kamalika Chaudhuri, Stefanie Jegelka, Le~Song, Csaba
  Szepesv{\'{a}}ri, Gang Niu, and Sivan Sabato (eds.), \emph{ICML}, 2022.

\bibitem[Huang et~al.(2019)Huang, Vaswani, Uszkoreit, Simon, Hawthorne,
  Shazeer, Dai, Hoffman, Dinculescu, and Eck]{music_transformer}
Cheng{-}Zhi~Anna Huang, Ashish Vaswani, Jakob Uszkoreit, Ian Simon, Curtis
  Hawthorne, Noam Shazeer, Andrew~M. Dai, Matthew~D. Hoffman, Monica
  Dinculescu, and Douglas Eck.
\newblock Music transformer: Generating music with long-term structure.
\newblock In \emph{ICLR}, 2019.

\bibitem[Lam et~al.(2022)Lam, Wang, Su, and Yu]{audio_gen_diffusion}
Max W.~Y. Lam, Jun Wang, Dan Su, and Dong Yu.
\newblock {BDDM:} bilateral denoising diffusion models for fast and
  high-quality speech synthesis.
\newblock In \emph{ICLR}, 2022.

\bibitem[Lee et~al.(2019)Lee, Lee, Kim, Kosiorek, Choi, and
  Teh]{settransformer}
Juho Lee, Yoonho Lee, Jungtaek Kim, Adam~R. Kosiorek, Seungjin Choi, and
  Yee~Whye Teh.
\newblock Set transformer: {A} framework for attention-based
  permutation-invariant neural networks.
\newblock In \emph{ICML}, 2019.

\bibitem[Liu et~al.(2020)Liu, Zou, Deng, Zuo, Lai, Ma, Liu, and
  Wang]{liu2020scenesketcher}
Fang Liu, Changqing Zou, Xiaoming Deng, Ran Zuo, Yu{-}Kun Lai, Cuixia Ma,
  Yong{-}Jin Liu, and Hongan Wang.
\newblock Scenesketcher: Fine-grained image retrieval with scene sketches.
\newblock In \emph{ECCV}, 2020.

\bibitem[Liu et~al.(2022)Liu, Ren, Lin, and Zhao]{plms_sampler}
Luping Liu, Yi~Ren, Zhijie Lin, and Zhou Zhao.
\newblock Pseudo numerical methods for diffusion models on manifolds.
\newblock In \emph{ICLR}, 2022.

\bibitem[Lopes et~al.(2019)Lopes, Ha, Eck, and Shlens]{fontgen_iccv}
Raphael~Gontijo Lopes, David Ha, Douglas Eck, and Jonathon Shlens.
\newblock A learned representation for scalable vector graphics.
\newblock In \emph{ICCV}, 2019.

\bibitem[Loshchilov \& Hutter(2019)Loshchilov and Hutter]{adamw}
Ilya Loshchilov and Frank Hutter.
\newblock Decoupled weight decay regularization.
\newblock In \emph{ICLR}, 2019.

\bibitem[Luhman \& Luhman(2020)Luhman and Luhman]{diffusion_handwriting}
Troy Luhman and Eric Luhman.
\newblock Diffusion models for handwriting generation.
\newblock \emph{CoRR}, abs/2011.06704, 2020.

\bibitem[Luo \& Hu(2021{\natexlab{a}})Luo and Hu]{shitong_denoising_score}
Shitong Luo and Wei Hu.
\newblock Score-based point cloud denoising.
\newblock In \emph{ICCV}, 2021{\natexlab{a}}.

\bibitem[Luo \& Hu(2021{\natexlab{b}})Luo and Hu]{shitong_diff3d}
Shitong Luo and Wei Hu.
\newblock Diffusion probabilistic models for 3d point cloud generation.
\newblock In \emph{CVPR}, 2021{\natexlab{b}}.

\bibitem[Muhammad et~al.(2019)Muhammad, Yang, Hospedales, Xiang, and
  Song]{umar_abstraction}
Umar~Riaz Muhammad, Yongxin Yang, Timothy~M Hospedales, Tao Xiang, and Yi-Zhe
  Song.
\newblock Goal-driven sequential data abstraction.
\newblock In \emph{ICCV}, 2019.

\bibitem[Nichol \& Dhariwal(2021)Nichol and Dhariwal]{iddpm}
Alexander~Quinn Nichol and Prafulla Dhariwal.
\newblock Improved denoising diffusion probabilistic models.
\newblock In \emph{ICML}, 2021.

\bibitem[Nichol et~al.(2022)Nichol, Dhariwal, Ramesh, Shyam, Mishkin, McGrew,
  Sutskever, and Chen]{glide_openai}
Alexander~Quinn Nichol, Prafulla Dhariwal, Aditya Ramesh, Pranav Shyam, Pamela
  Mishkin, Bob McGrew, Ilya Sutskever, and Mark Chen.
\newblock {GLIDE:} towards photorealistic image generation and editing with
  text-guided diffusion models.
\newblock In Kamalika Chaudhuri, Stefanie Jegelka, Le~Song, Csaba
  Szepesv{\'{a}}ri, Gang Niu, and Sivan Sabato (eds.), \emph{ICML}, 2022.

\bibitem[Pang et~al.(2019)Pang, Li, Yang, Zhang, Hospedales, Xiang, and
  Song]{kaiyue19sbir}
K.~Pang, K.~Li, Y.~Yang, H.~Zhang, T.~M. Hospedales, T.~Xiang, and Y.~{-}Z.
  Song.
\newblock Generalising fine-grained sketch-based image retrieval.
\newblock In \emph{CVPR}, 2019.

\bibitem[Qi et~al.(2017)Qi, Su, Mo, and Guibas]{pointnet}
Charles~Ruizhongtai Qi, Hao Su, Kaichun Mo, and Leonidas~J. Guibas.
\newblock Pointnet: Deep learning on point sets for 3d classification and
  segmentation.
\newblock In \emph{CVPR}, 2017.

\bibitem[Ramesh et~al.(2021)Ramesh, Pavlov, Goh, Gray, Voss, Radford, Chen, and
  Sutskever]{dalle_openai}
Aditya Ramesh, Mikhail Pavlov, Gabriel Goh, Scott Gray, Chelsea Voss, Alec
  Radford, Mark Chen, and Ilya Sutskever.
\newblock Zero-shot text-to-image generation.
\newblock In Marina Meila and Tong Zhang (eds.), \emph{ICML}, 2021.

\bibitem[Ribeiro et~al.(2020)Ribeiro, Bui, Collomosse, and Ponti]{sketchformer}
Leo Sampaio~Ferraz Ribeiro, Tu~Bui, John~P. Collomosse, and Moacir Ponti.
\newblock Sketchformer: Transformer-based representation for sketched
  structure.
\newblock In \emph{CVPR}, 2020.

\bibitem[Rombach et~al.(2021)Rombach, Blattmann, Lorenz, Esser, and
  Ommer]{latent_diffusion}
Robin Rombach, Andreas Blattmann, Dominik Lorenz, Patrick Esser, and
  Bj{\"{o}}rn Ommer.
\newblock High-resolution image synthesis with latent diffusion models.
\newblock \emph{CVPR}, 2021.

\bibitem[Sohl{-}Dickstein et~al.(2015)Sohl{-}Dickstein, Weiss, Maheswaranathan,
  and Ganguli]{diffusionmodel}
Jascha Sohl{-}Dickstein, Eric~A. Weiss, Niru Maheswaranathan, and Surya
  Ganguli.
\newblock Deep unsupervised learning using nonequilibrium thermodynamics.
\newblock In \emph{ICML}, 2015.

\bibitem[Song et~al.(2021{\natexlab{a}})Song, Meng, and Ermon]{ddim}
Jiaming Song, Chenlin Meng, and Stefano Ermon.
\newblock Denoising diffusion implicit models.
\newblock In \emph{ICLR}, 2021{\natexlab{a}}.

\bibitem[Song et~al.(2021{\natexlab{b}})Song, Sohl{-}Dickstein, Kingma, Kumar,
  Ermon, and Poole]{diffusion_SDE_yangsong}
Yang Song, Jascha Sohl{-}Dickstein, Diederik~P. Kingma, Abhishek Kumar, Stefano
  Ermon, and Ben Poole.
\newblock Score-based generative modeling through stochastic differential
  equations.
\newblock In \emph{ICLR}, 2021{\natexlab{b}}.

\bibitem[Srivastava et~al.(2015)Srivastava, Mansimov, and
  Salakhudinov]{srivastava2015unsupervised}
Nitish Srivastava, Elman Mansimov, and Ruslan Salakhudinov.
\newblock Unsupervised learning of video representations using lstms.
\newblock In \emph{ICML}, 2015.

\bibitem[Su et~al.(2020)Su, Qi, Pang, Yang, and Song]{sketchhealing}
Guoyao Su, Yonggang Qi, Kaiyue Pang, Jie Yang, and Yi{-}Zhe Song.
\newblock Sketchhealer: {A} graph-to-sequence network for recreating partial
  human sketches.
\newblock In \emph{BMVC}, 2020.

\bibitem[Tashiro et~al.(2021)Tashiro, Song, Song, and Ermon]{csdi_imputation}
Yusuke Tashiro, Jiaming Song, Yang Song, and Stefano Ermon.
\newblock {CSDI:} conditional score-based diffusion models for probabilistic
  time series imputation.
\newblock In \emph{NeurIPS}, 2021.

\bibitem[van~den Oord et~al.(2016)van~den Oord, Dieleman, Zen, Simonyan,
  Vinyals, Graves, Kalchbrenner, Senior, and Kavukcuoglu]{wavenet}
A{\"{a}}ron van~den Oord, Sander Dieleman, Heiga Zen, Karen Simonyan, Oriol
  Vinyals, Alex Graves, Nal Kalchbrenner, Andrew~W. Senior, and Koray
  Kavukcuoglu.
\newblock Wavenet: {A} generative model for raw audio.
\newblock In \emph{ISCA}, 2016.

\bibitem[Vaswani et~al.(2017)Vaswani, Shazeer, Parmar, Uszkoreit, Jones, Gomez,
  Kaiser, and Polosukhin]{transformer}
Ashish Vaswani, Noam Shazeer, Niki Parmar, Jakob Uszkoreit, Llion Jones,
  Aidan~N. Gomez, Lukasz Kaiser, and Illia Polosukhin.
\newblock Attention is all you need.
\newblock In \emph{NeurIPS}, 2017.

\bibitem[Wang et~al.(2020)Wang, Lin, Li, Wu, Cai, Luo, and Wang]{wang2020multi}
Fei Wang, Shujin Lin, Hanhui Li, Hefeng Wu, Tie Cai, Xiaonan Luo, and Ruomei
  Wang.
\newblock Multi-column point-cnn for sketch segmentation.
\newblock \emph{Neurocomputing}, 2020.

\bibitem[Xu et~al.(2022)Xu, Yu, Song, Shi, Ermon, and Tang]{geodiff}
Minkai Xu, Lantao Yu, Yang Song, Chence Shi, Stefano Ermon, and Jian Tang.
\newblock Geodiff: {A} geometric diffusion model for molecular conformation
  generation.
\newblock In \emph{ICLR}, 2022.

\bibitem[Yang et~al.(2021)Yang, Zhuang, Fu, Wei, Zhou, and
  Zheng]{yang2021sketchgnn}
Lumin Yang, Jiajie Zhuang, Hongbo Fu, Xiangzhi Wei, Kun Zhou, and Youyi Zheng.
\newblock Sketch{GNN}: Semantic sketch segmentation with graph neural networks.
\newblock \emph{ACM Transactions on Graphics (TOG)}, 2021.

\bibitem[Yu et~al.(2015)Yu, Yang, Song, Xiang, and Hospedales]{sketchanet1}
Qian Yu, Yongxin Yang, Yi-Zhe Song, Tao Xiang, and Timothy Hospedales.
\newblock Sketch-a-net that beats humans.
\newblock In \emph{BMVC}, 2015.

\bibitem[Yu et~al.(2017)Yu, Yang, Liu, Song, Xiang, and
  Hospedales]{sketchanet2}
Qian Yu, Yongxin Yang, Feng Liu, Yi-Zhe Song, Tao Xiang, and Timothy
  Hospedales.
\newblock Sketch-a-net: A deep neural network that beats humans.
\newblock \emph{IJCV}, 122:\penalty0 411–425, 2017.

\bibitem[Zaheer et~al.(2017)Zaheer, Kottur, Ravanbakhsh, P{\'{o}}czos,
  Salakhutdinov, and Smola]{deepsets}
Manzil Zaheer, Satwik Kottur, Siamak Ravanbakhsh, Barnab{\'{a}}s P{\'{o}}czos,
  Ruslan Salakhutdinov, and Alexander~J. Smola.
\newblock Deep sets.
\newblock In \emph{NeurIPS}, 2017.

\end{thebibliography}
